# Web Service QoS Prediction via Extended Canonical Polyadic-based Tensor Network

Qu Wang, and Hao Wu

*Abstract*—Today, numerous web services with similar functionalities are available on the Internet. Users often evaluate the Quality of Service (QoS) to choose the best option among them. Predicting the QoS values of these web services is a significant challenge in the field of web services. A Canonical Polyadic (CP)-based tensor network model has proven to be efficient for predicting dynamic QoS data. However, current CP-based tensor network models do not consider the correlation of users and services in the low-dimensional latent feature space, thereby limiting model's prediction capability. To tackle this issue, this paper proposes an Extended Canonical polyadic-based Tensor Network (ECTN) model. It models the correlation of users and services via building a relation dimension between user feature and service feature in low-dimensional space, and then designs an extended CP decomposition structure to improve prediction accuracy. Experiments are conducted on two public dynamic QoS data, and the results show that compared with state-of-the-art QoS prediction models, the ECTN obtains higher prediction accuracy.

## I. Introduction

With the development and popularization of the Internet, the demand for network Quality of Service (QoS) has become increasingly urgent [1-8]. QoS is a key indicator for measuring network performance and service quality, and is crucial to ensuring user experience and meeting different application needs. However, in actual network environments, QoS data are often incomplete or inaccurate due to various reasons [9, 10], such as network congestion, equipment failure, data collection errors, etc. In this context, QoS prediction by utilizing historical data has become an important technology, which can help network administrators and service providers identify potential problems in advance and take appropriate measures to maintain and optimize network services [11]. Thus, QoS prediction is important in network management and operation, and helps to improve the stability and reliability of the network [12-19].

The earliest QoS prediction methods usually rely on static data characteristics and ignore the dynamic characteristics of QoS data, which may lead to inaccurate or insufficient adaptability of prediction results [5, 20, 21]. Dynamic QoS prediction methods can better capture real-time changes in QoS data. In recent years, many methods for dynamic QoS prediction have been proposed [8, 22-29], among which the tensor decomposition -based models have received more attention due to its excellent performance. It models dynamic QoS data as a third-order tensor, which preserves the spatial structure and temporal pattern of QoS data well [ 2, 6, 8, 9, 11, 30-38]. However, most of the existing QoS prediction models use Canonical Polyadic (CP)-based tensor network model, and their representation ability is limited by CP decomposition structure because the correlation between users and services in low-dimensional latent feature space are not considered, making it difficult to obtain higher prediction accuracy. Tensor network is a powerful tool for representing and processing multidimensional data, which shows good definability and interpretability [39-44]. Therefore, can we design a new CP decomposition structure to build a tensor network that is able to model the correlation of users and services in the low-dimensional latent feature space?

In order to verify the above idea, this paper innovatively proposes an Extended Canonical polyadic-based Tensor Network (ECTN) model for dynamic QoS prediction. Firstly, the correlation of users and services is modeled by building a relation dimension between user feature and service feature in low-dimensional space, and an extended CP decomposition structure is designed to build tensor network. Secondly, a data density-oriented idea is adopted to construct learning objectives on incomplete tensors to reduce computational and storage costs. Then, a nonnegative multiplication updates on incomplete tensors (NMU-IT) algorithm is designed for ECTN, which ensures the non-negativity of QoS prediction values during training. This paper's main contributions are as follows:

a) A ECTN model. It is able to accurately predict dynamic QoS data by modeling the correlation of users and services in the low-dimensional latent feature space.

b) Detailed design of an NMU-IT algorithm for ECTN. It can not only achieve rapid convergence but also maintain the non-negativity of QoS prediction values during the training process, and experiments have proven its effectiveness.

c) Experiments are conducted on two public QoS datasets. Compared with other state-of-the-art models, ECTN has higher prediction accuracy.

Q. Wang, and H. Wu are with the College of Computer and Information Science, Southwest University, Chongqing 400715, China, (email: wangquff@gamil.com, haowuf@gamil.com).

The remainder of this article is organized as follows: Section II reviews related work, Section III introduces the prerequisites, Section IV designs the ECTN model, Section V conducts experiments and analysis, and Section VI concludes this paper.

## II. RELATED WORK

### A. Static QoS Prediction

Static QoS prediction models usually model QoS data as a user-service matrix, and then extract its features to predict unknown values. Wu *et al*. [45] propose a data feature-aware latent factor model, which utilizes data feature-aware and density clustering techniques to improve the accuracy of QoS prediction. Li *et al*. [46] build a topology-aware neural network that improves the accuracy of QoS prediction by comprehensively considering the characteristics of users, services, and nodes on the communication path. Xia *et al*. [47] extract and learn implicit and explicit features and their interactions in multi-source QoS data by combining matrix factorization, neural network and convolutional neural network techniques. Lian *et al*. [48] propose a personalized service QoS prediction method from the perspective of multi-task learning, through feature selection of multi-expert decision-making and self-attention mechanisms, and multi-step model training without weight configuration. Wu *et al*. [49] design the Reputation Integrated Graph Convolution Network to improve the accuracy and robustness of QoS prediction by combining user reputation extraction, multi-source feature extraction and graph convolution network.

The main disadvantage of static QoS prediction models is that they cannot adapt to real-time dynamic changes in network environment and service conditions, resulting in insufficient accuracy and robustness of prediction results.

### B. Dynamic QoS Prediction

The advantage of the dynamic QoS prediction model is that it can adapt to real-time changes in the network environment and service conditions and provide more accurate and robust service quality predictions. Yuan *et al*. [50] propose a QoS estimator incorporating latent factor analysis with Kalman filtering for accurately modeling latent temporal patterns in dynamic QoS data. Luo *et al*. [51] design an tensor learning model based on CP decomposition for dynamic QoS prediction, and design linear bias for CP decomposition. Ye *et al*. [52] use Cauchy loss to establish the error between observed values and predicted values to obtain more robust QoS prediction results. Su *et al*. [53] build a context-aware QoS prediction model using CP decomposition, which is updated using gradient descent and least squares. Gao *et al*. [54] propose a dynamic reconstruction method under the cloud edge computing model by extending the concept of service quality, using LSTM network to predict service stability and calculate service invocation costs.

Although the above models can predict QoS values, their performance is limited by the basic model of tensor CP decomposition. This paper uses block term decomposition to build a dynamic QoS prediction model, which can capture the complex correlation between users and matrices to obtain more accurate prediction results.

## III. PRELIMINARIES

### A. A Dynamic QoS Data Tensor

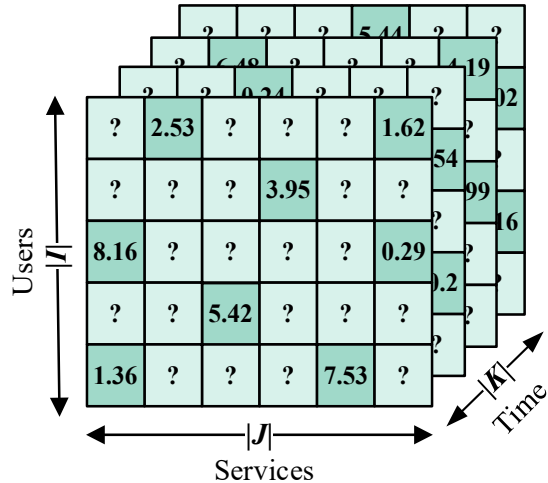

Fig. 1. A dynamic QoS data tensor.

Dynamic QoS data refers to service quality-related indicators that change over time [55-60], such as response time and throughput, etc. The indicators can reflect the performance of network services or applications at different points in time. As shown in Fig. 1, a dynamic QoS tensor $\mathbf{Y} \in \mathbb{R}^{|I| \times |J| \times |K|}$ is constructed by sequentially arranging the user-service matrix at different times, where $I$ represents the set of users, $J$ represents the set of services, and $K$ represents the time set. Since it is impossible for users to access all services at the same time, a QoS tensor Y is highly incomplete, that is, the known entries set $\Lambda$ is far smaller than the size of the tensor $|I| \times |J| \times |K|$.

## B. Tensor Network Notations

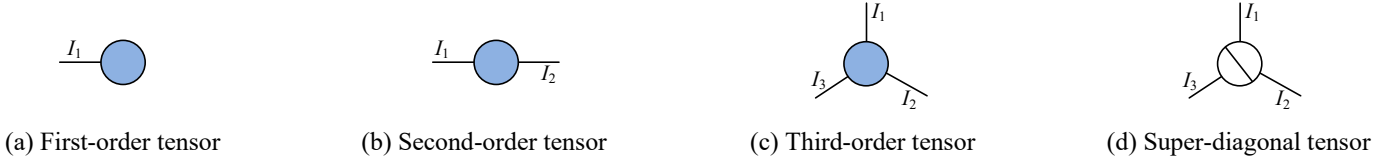

(a) First-order tensor  (b) Second-order tensor  (c) Third-order tensor  (d) Super-diagonal tensor

Fig. 2. Tensor form in tensor network.

In tensor network, each tensor is represented by a node with an edge. As shown in Fig. 2, a node with an edge represents a first-order tensor, i.e., a vector $y$. A node with two edges represents a second-order tensor, i.e., a matrix Y. A node with three edges represents a third-order tensor, i.e., a tensor **Y**. A super-diagonal tensor is a 3rd-order tensor in which only the elements on its super-diagonal are constant 1.

## IV. A ECTN MODEL

### A. Objective Function

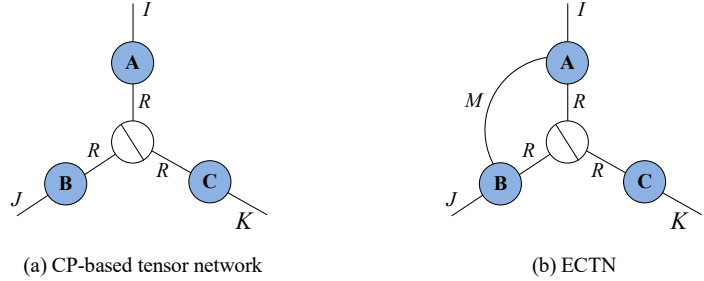

(a) CP-based tensor network  (b) ECTN

Fig. 3. Graph illustration from CP-based tensor network to ECTN.

According to previous research [61-66], CP decomposition usually decomposes a third-order tensor into the sum of $R$ rank-one tensors, and each rank tensor is obtained by the column vector outer product of the factor matrix. Therefore, CP decomposition can be described by the following formula:

$$\mathbf{Y} \approx \sum_{r=1}^{R} \boldsymbol{a}_r \circ \boldsymbol{b}_r \circ \boldsymbol{c}_r. \tag{1}$$

As shown in Fig. 3(a), in the tensor network, CP decomposition is represented in the form of a super-diagonal tensor and three matrices, and each factor matrix is independently connected to an edge of the super-diagonal tensor. If the $I$ and $J$ dimensions of a third-order tensor contain more complex linear structures or have some similarities, the CP-based tensor network cannot accurately represent it. As shown in Fig. 3(b), ECTN connects nodes A and B in a CP-based tensor network, which is not just a simple connection operation, but also multi-dimensional linear expansion of matrices A and B into third-order tensors. As shown in Fig. 4, The essence of ECTN is to multidimensionally linearly expand the vectors $a_r \in \mathbb{R}^{|I| \times 1}$ and $b_r \in \mathbb{R}^{|J| \times 1}$ in (1) into matrices $A_r \in \mathbb{R}^{|I| \times M}$ and $B_r \in \mathbb{R}^{|J| \times M}$, where $M$ represents the expanded dimension. Therefore, a third-order tensor $\mathbf{Y} \in \mathbb{R}^{|I| \times |J| \times |K|}$ representation of ECTN is obtained:

$$\mathbf{Y} \approx \sum_{r=1}^{R} (A_r B_r^T) \circ \boldsymbol{c}_r. \tag{2}$$

In order to establish a fine-grained objective function, the following derivation is first performed. Define the intermediate state matrix $Z_r \in \mathbb{R}^{|I| \times |J|}$ as follows:

$$Z_r = A_r B_r^T. \tag{3}$$

Then, each element in $Z_r$ is calculated as:

$$z_{ij,r} = \sum_{m=1}^{M} a_{im,r} b_{jm,r}, \tag{4}$$

where $z_{ij,r}$ represents the value of the $r$-th matrix $Z_r$ at the position $(i, j)$, and the same goes for $a_{im,r}$ and $b_{jm,r}$. In order to facilitate understanding and calculation, define tensor $\mathbf{Z} \in \mathbb{R}^{|I| \times |J| \times R}$, which represents $(Z_1,...,Z_r)$, tensor $\mathbf{A} \in \mathbb{R}^{|I| \times M \times R}$, which represents $(A_1,...,A_r)$, tensor $\mathbf{B} \in \mathbb{R}^{|J| \times M \times R}$, which represents $(B_1,...,B_r)$, and matrix $C \in \mathbb{R}^{|K| \times R}$, which represents $(c_1,...,c_r)$. This setting also verifies the representation of ECTN in Fig. 3(b), and is also related to the tensor decomposition structure of ECTN in Fig. 4. According to the above definition and through formula (4), the calculation formula of a single element in tensor $\hat{\mathbf{Y}}$ is obtained:

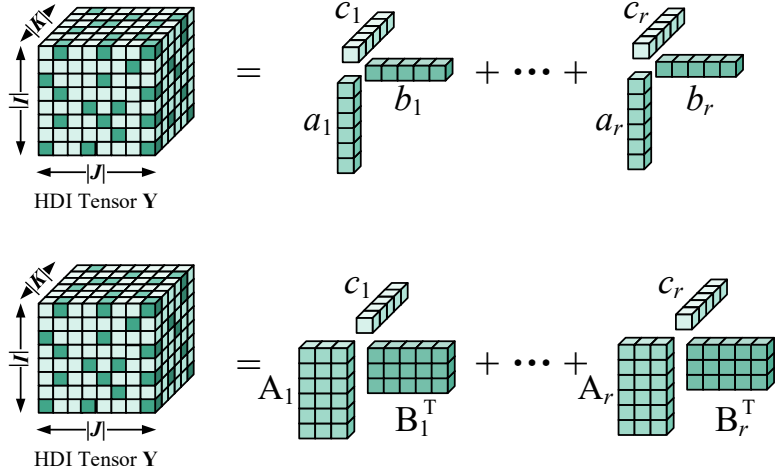

Fig. 4. Tensor decomposition structure of CP and ECTN. (The top of this figure is CP, and the bottom is ECTN)

$$\hat{y}_{ijk} = \sum_{r=1}^{R} z_{ijr} c_{kr} = \sum_{r=1}^{R} \sum_{m=1}^{M} a_{imr} b_{jmr} c_{kr}. \tag{5}$$

In order to obtain the latent factors **A**, **B**, C, the widely used Euclidean distance is adopted in this work to measure the difference between **Y** and **Ŷ** [67-70]. From this, it is given:

$$\varepsilon = \left\| \mathbf{Y} - \hat{\mathbf{Y}} \right\|_{\mathrm{F}}^{2} = \sum_{i=1}^{|I|} \sum_{j=1}^{|J|} \sum_{k=1}^{|K|} \left( y_{ijk} - \hat{y}_{ijk} \right)^{2}. \tag{6}$$

As analyzed in section III part A and shown in Fig. 1, most of the data in a QoS tensor **Y** is unknown. Therefore, the objective function should be defined on the known item set Λ to reduce the computational cost of the model. Then, by substituting (5) into (6) to get:

$$\varepsilon = \sum_{y_{ijk} \in \Lambda} \left( y_{ijk} - \sum_{r=1}^{R} \sum_{m=1}^{M} a_{imr} b_{jmr} c_{kr} \right)^{2}. \tag{7}$$

At this point, a general objective function of ECTN learning model is obtained. But for another problem, dynamic QoS data often has data fluctuations, and its needs to be modeled into the objective function to enhance the learning ability of the model. According to previous research, incorporating the linear bias of each dimension into the learning objective can model QoS data fluctuations well. Therefore, incorporating three linear bias vectors $\boldsymbol{d} \in \mathbb{R}^{1 \times |I|}$, $\boldsymbol{e} \in \mathbb{R}^{1 \times |J|}$, and $\boldsymbol{f} \in \mathbb{R}^{1 \times |K|}$ in (5) gives:

$$\hat{y}_{ijk} = \sum_{r=1}^{R} \sum_{m=1}^{M} a_{imr} b_{jmr} c_{kr} + d_{i} + e_{j} + f_{k}, \tag{8}$$

where $\boldsymbol{d}$, $\boldsymbol{e}$, and $\boldsymbol{f}$ represent the linear bias of the user, service, and time dimensions respectively, $d_i$ represents the $i$-th element of the vector $\boldsymbol{d}$, and $e_j$ and $f_k$ are similar. Therefore, a biased objective function is regained:

$$\varepsilon = \sum_{y_{ijk} \in \Lambda} \left( y_{ijk} - \sum_{r=1}^{R} \sum_{m=1}^{M} a_{imr} b_{jmr} c_{kr} + d_{i} + e_{j} + f_{k} \right)^{2}. \tag{9}$$

Since (9) is ill-posed and the data distribution in QoS tensor **Y** is unbalanced. Therefore, it is crucial to introduce regularization terms into the objective function, which can not only help solve ill-posed problems but also prevent model overfitting [71-75]. Since $L_2$ regularization can make the model more robust to outliers, it is included in (9):

$$\varepsilon = \sum_{y_{ijk} \in \Lambda} \left( \left( y_{ijk} - \sum_{r=1}^{R} \sum_{m=1}^{M} a_{imr} b_{jmr} c_{kr} + d_{i} + e_{j} + f_{k} \right)^{2} + \lambda \left( \sum_{r=1}^{R} \sum_{m=1}^{M} \left( a_{imr}^{2} + b_{jmr}^{2} \right) + \sum_{r=1}^{R} c_{kr}^{2} + d_{i}^{2} + e_{j}^{2} + f_{k}^{2} \right) \right),$$

$$s.t.\ i \in I, j \in J, k \in K, r \in \{1,...,R\}, m \in \{1,...,M\}: a_{imr} \geq 0, b_{jmr} \geq 0, c_{kr} \geq 0, d_{i} \geq 0, e_{j} \geq 0, f_{k} \geq 0. \tag{10}$$

where $\lambda$ is the regularization coefficient. Since the values in the QoS tensor are all nonnegative, for example, the response time cannot be negative. Therefore, a nonnegative constraint on the latent factor is imposed in (10) to ensure that the QoS predicted value is nonnegative.

## B. Parameters Learning Scheme

A nonnegative multiplication updates on incomplete tensors (NMU-IT) algorithm is usually used to solve tensor learning models [70, 72, 75-77], which sets an adaptive learning rate during the gradient descent process to ensure the non-negativity of the latent factors. Since the derivation process of **A** and **B** is similar, the following part only shows that of **A**, and the same is true for *d*, *e*, and *f*. According to the principle of NMU-IT, first perform the partial derivative of each latent factor in (10) to obtain:

$$\begin{cases} a_{imr} \leftarrow a_{imr} - \eta_{imr} \sum_{y_{ijk} \in \Lambda(i)} \left( \lambda a_{imr} - (y_{ijk} - \hat{y}_{ijk}) b_{jmr} c_{kr} \right), \\ c_{kr} \leftarrow c_{kr} - \eta_{kr} \sum_{y_{ijk} \in \Lambda(k)} \left( \lambda c_{kr} - (y_{ijk} - \hat{y}_{ijk}) \sum_{m=1}^{M} a_{imr} b_{jmr} \right), \\ d_i \leftarrow d_i - \eta_i \sum_{y_{ijk} \in \Lambda(i)} \left( \lambda d_i - (y_{ijk} - \hat{y}_{ijk}) \right), \end{cases} \quad (11)$$

where $\eta_{imr}$, $\eta_{kr}$, and $\eta_i$ respectively represent the learning rate of the corresponding latent factor. Note that the negative term in (11) will cause the latent factor to become negative during training. In order to strictly guarantee nonnegativity, the learning rate of each latent factor is set as follows to eliminate these negative terms:

$$\begin{cases} \eta_{imr} = a_{imr} \Big/ \sum_{y_{ijk} \in \Lambda(i)} \left( \lambda a_{imr} + \hat{y}_{ijk} b_{jmr} c_{kr} \right), \\ \eta_{kr} = c_{kr} \Big/ \sum_{y_{ijk} \in \Lambda(k)} \left( \lambda c_{kr} + \hat{y}_{ijk} \sum_{m=1}^{M} a_{imr} b_{jmr} \right), \\ \eta_i = d_i \Big/ \sum_{y_{ijk} \in \Lambda(i)} \left( \lambda d_i + \hat{y}_{ijk} \right). \end{cases} \quad (12)$$

Then, (12) is brought into (11), and simplified to obtain the following update formula:

$$\begin{cases} a_{imr} \leftarrow a_{imr} \dfrac{\sum_{y_{ijk} \in \Lambda(i)} y_{ijk} b_{jmr} c_{kr}}{\sum_{y_{ijk} \in \Lambda(i)} \left( \lambda a_{imr} + \hat{y}_{ijk} b_{jmr} c_{kr} \right)}, \\ c_{kr} \leftarrow c_{kr} \dfrac{\sum_{y_{ijk} \in \Lambda(k)} y_{ijk} \sum_{m=1}^{M} a_{imr} b_{jmr}}{\sum_{y_{ijk} \in \Lambda(k)} \left( \lambda c_{kr} + \hat{y}_{ijk} \sum_{m=1}^{M} a_{imr} b_{jmr} \right)}, \\ d_i \leftarrow d_i \dfrac{\sum_{y_{ijk} \in \Lambda(i)} y_{ijk}}{\sum_{y_{ijk} \in \Lambda(i)} \lambda d_i + \hat{y}_{ijk}}. \end{cases} \quad (13)$$

Note that in (13), just initializing each latent factor to be nonnegative can ensure its nonnegativity during the training process. And NMU-IT, an algorithm that implicitly relies on gradients, does not require additional time cost for adjusting the learning rate, and its time cost saving is obvious. Through the above settings, we get a parameter training scheme based on NMU-IT algorithm.

---
**Algorithm 1. ECTN**
**Operation**
1: **Input**: |*I*|, |*J*|, |*K*|, *R*, *M*, Λ
2: **Initialization A**, **B**, C, *d*, *e*, *f* with small random numbers
3: **While** not convergence **do**
4:   **for** each $y_{ijk}$ in Λ
5:     compute $\hat{y}_{ijk}$ based on (8)
6:     update **A** and **B** based on (13)
7:     update C based on (13)
8:     update *d*, *e* and *f* based on (13)
9:   **end for**
10: **end while**
11: **Output**: **A**, **B**, C, *d*, *e*, *f*
---

## C. Algorithm Design and Analysis

Based on the above analysis and derivation, Algorithm 1 ECTN is given. The computational complexity $C_{ECTN}$ mainly lies in the initialization process $C_1$ and training update process $C_2$, which are given below:

$$C_1 = \Theta\left((|I|+|J|)\times(R\times M+1)+|K|\times(R+1)\right),$$
$$C_2 = \Theta\left((|I|+|J|+|\Lambda|)\times(R\times M+1)+|K|\times(R+1)\right).$$
(14)

Among them, $C_1$ includes initialization latent factors **A**, **B**, C and bias vectors *d*, *e*, *f*, and $C_2$ represents the computational complexity of its update process. Then, $C_{ECTN}$ is given:

$$C_{ECTN} = C_1 + C_2 \approx |\Lambda|\times R\times M.$$
(15)

Note that constant coefficients and low-order terms are omitted in $C_{ECTN}$. It can be found that the computational complexity of ECTN has a linear relationship with the size of known elements set $\Lambda$ in the tensor. Then calculate the space complexity of ECTN below,

$$S_{ECTN} = \Theta\left(|\Lambda|+(|I|+|J|)\times(R\times M+1)+|K|\times(R+1)\right)$$
$$\approx \Theta\left(|\Lambda|+(|I|+|J|)\times R\times M\right).$$
(16)

The space complexity of ECTN is mainly related to the size of the latent factor and the size of known entries set $\Lambda$ in the QoS tensor.

## V. EXPERIMENTS

### A. Experimental Setup

TABLE I. DATASETS

| Datasets | | $|\Omega|:|\Psi|:|\Phi|$ | $|\Omega|+|\Psi|+|\Phi|$ | User Count | Service Count | Time Count |
|---|---|---|---|---|---|---|
| D1 | D1.1 | 1%:9%:90% | 30,171,491 | 142 | 4500 | 64 |
|    | D1.2 | 5%:5%:90% | 30,171,491 | 142 | 4500 | 64 |
| D2 | D2.1 | 1%:9%:90% | 25,652,333 | 142 | 4500 | 64 |
|    | D2.2 | 5%:5%:90% | 25,652,333 | 142 | 4500 | 64 |

*Datasets*: This work uses two publicly available dynamic QoS data, D1 (response time) and D2 (throughput) [78], which record response time and throughput between users and services, respectively. As shown in Table I, each dataset contains interactions between 142 users and 4500 services in 64 time periods. In order to simulate different data loss situations, D1 and D2 are divided into training set $\Omega$, verification set $\Psi$, and test set $\Phi$ according to two different proportions, and then four different sub-datasets are obtained respectively.

*Experiment details*: In order to fairly verify the performance of the model, all models are implemented on i7 CPU using the JAVA language JDK1.8 environment. The model is considered to have converged when the loss in two consecutive rounds is less than the threshold $10^{-5}$ or the number of iteration rounds exceeds 1000. And each sub-data set is randomly divided 10 times, and the final result is the average of the 10 results.

*Evaluation Metrics*: According to previous research [79-81], Root Mean Square Error (RMSE) and Mean Absolute Error (MAE) are commonly used statistical indicators to measure the accuracy of prediction models, especially in the fields of regression analysis and forecasting. The smaller the RMSE and MAE, the higher the accuracy of the prediction. The calculation formula is as follows:

$$\text{RMSE} = \sqrt{\frac{1}{|\Phi|}\sum_{y_{ijk}\in\Omega}(y_{ijk}-\hat{y}_{ijk})^2}, \text{MAE} = \frac{1}{|\Phi|}\sum_{y_{ijk}\in\Omega}|y_{ijk}-\hat{y}_{ijk}|.$$

*Baselines*: To verify the performance of the proposed ECTN, it is compared with the following state-of-the-art models:

- **M1**: A ECTN model proposed in this paper.
- **M2**: A QoS prediction model using the Cauchy loss to build losses to increase the model's robustness to outliers [52].
- **M3**: An integrated environment QoS prediction model based on CP decomposition using gradient descent and least squares optimization models [53].
- **M4**: A biased tensor learning model incorporates the idea of transfer learning for QoS prediction [65].
- **M5**: A time-aware QoS prediction model that splits dynamic QoS data into user-service slices and models temporal patterns into objective functions [82].

In order to make a fair comparison, the latent feature dimension $R$ or $M$ of all models is set to 5.

*Hyperparameter settings*: The hyperparameter that affects the performance of the ECTN model is the regularization coefficient λ. We search λ from the range [0.1-1.0] with a step size of 0.1, and finally decide to set λ=0.4 on D1.1 and D1.2, and set λ=0.8 on D2.1 and D2.2. For the hyperparameters of other baselines, they are fine-tuned from the settings in their papers.

## B. Experimental results and analysis

TABLE II.   RMSE AND MAE OF M1-M5 ON DATASETS

| Datasets | | M1 | M2 | M3 | M4 | M5 |
|---|---|---|---|---|---|---|
| D1.1 | RMSE | **3.2854** | 3.3557 | 3.4119 | 3.3866 | 5.0858 |
|  | MAE | **1.5707** | 1.5826 | 1.6759 | 1.6757 | 2.8785 |
| D1.2 | RMSE | **3.0483** | 3.0963 | 3.1385 | 3.0919 | 4.9605 |
|  | MAE | **1.4277** | 1.4485 | 1.4913 | 1.4487 | 2.8499 |
| D2.1 | RMSE | **30.8957** | 31.7075 | 37.3340 | 36.8164 | 48.7248 |
|  | MAE | **5.3708** | 5.4546 | 7.2519 | 6.6160 | 10.3816 |
| D2.2 | RMSE | **24.8050** | 25.5490 | 26.2637 | 26.5888 | 48.1808 |
|  | MAE | **4.2463** | 4.3556 | 4.7116 | 4.6798 | 10.0175 |
| F-Rank | | 1.0 | 2.1 | 3.9 | 3.0 | 5.0 |

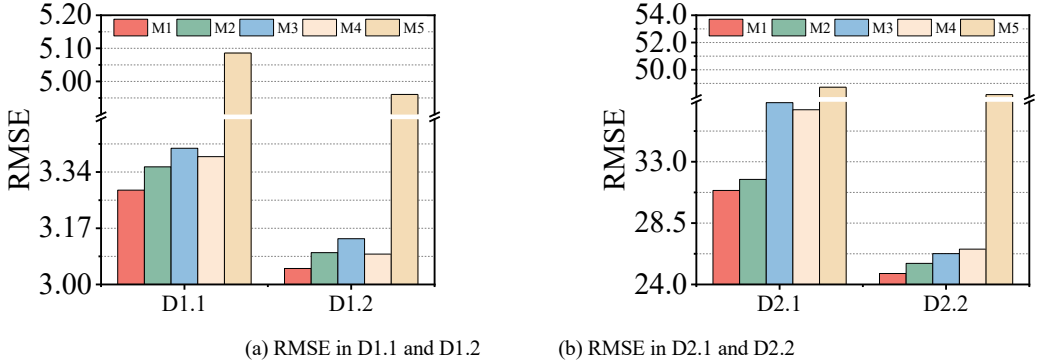

(a) RMSE in D1.1 and D1.2      (b) RMSE in D2.1 and D2.2

Fig. 5.   RMSE of all models on datasets

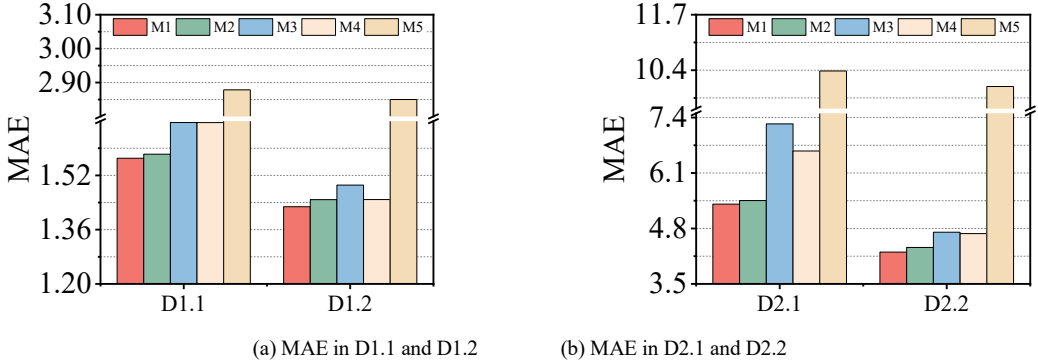

(a) MAE in D1.1 and D1.2      (b) MAE in D2.1 and D2.2

Fig. 6.   MAE of all models on datasets

Table 2 records the RMSE and MAE of M1-M5 on four cases. Fig. 5 and Fig. 6 show the RMSE and MAE of all models. From the results, we conduct the following analysis:

a) **The M1's RMSE is lower than other four models.** As shown in Table II and Fig. 5, the minimum RMSE achieved by M1 on D1.1 is 3.2854, which is 2.09% lower than the 3.3557 of M2, 3.71% lower than the 3.4119 of M3, 2.99% lower than the 3.3866 of M4, 35.40% lower than the 5.0858 of M5 respectively. Similar results are obtained on D1.2. In addition, the minimum RMSE achieved by M1 on D2.1 is 30.8957, which is 2.56% lower than the 31.7075 of M2, 17.25% lower than the 37.3340 of M3, 16.08 lower than the 36.8164 of M4, 36.59% lower than the 48.7248 of M5 respectively. The same is true for the results on D2.2. The above results prove that M1 has better QoS prediction accuracy. This is because M2-M4 are all QoS prediction models constructed using CP decomposition, while M1 is built on based ECTN, which is a multi-dimensional linear extension of CP decomposition in the user-service dimension, so better prediction accuracy is expected.

b) **The M1's MAE is lower than other four models.** As shown in Table II and Fig. 6, the minimum MAE achieved by M1 on D1.1 is 1.5707, while the other four models are 1.5826, 1.6759, 1.6757, and 2.8785 respectively. M1 is 0.75%, 6.28%, 6.27%, and 45.43% lower than other models respectively. The results on D1.2 are also the same. In addition the minimum MAE achieved by M1 on D2.1 is 5.3708, while the other four models are 5.4546, 7.2519, 6.6160, and 10.3816 respectively. M1 is 1.54%, 25.94%, 18.82%, and 48.27% lower than them respectively. Similar results can be found on D2.2. Based on the above analysis, it is found that the MAE of M1 decrease less than that of M2. This is because the Cauchy loss used by M2 is not sensitive to outliers in the QoS data set, but overall, the MAE of M1 is the lowest among all compared models.

c) **The M1's prediction accuracy is statistically confirmed.** The Friedman test is a non-parametric statistical test method commonly use to compare algorithm performance or experimental results under multiple processing conditions to determine whether there are significant differences between them. Its test results are reflected in F-Rank. In this work, the smaller the F-Rank, the higher the QoS prediction accuracy. According to the value of F-Rank, it can more intuitively judge the performance of the model. As shown in Table II, the F-Rank of M1 is 1.0, M2 is 2.1, M3 is 3.9, M4 is 3.0, and M5 is 5.0. Through statistical testing methods, it is confirmed that M1 has the best prediction accuracy among peer models. This is consistent with the above experimental result analysis.

## VI. CONCLUSION

In order to improve the accuracy of QoS prediction, this paper proposes an Extended Canonical polyadic-based Tensor Network (ECTN) model. This model establishes the connection between user and service dimensions based on the CP-based tensor network and performs multi-dimensional linear expansion in these dimensions. After deriving the calculation method of the ECTN model in detail, a nonnegative update algorithm is designed to optimize the model. Experimental results show that ECTN can predict unknown QoS data more accurately. ECTN has good scalability and is not limited to QoS prediction. In future work, we consider using different learning objectives to establish ECTN and use it in other representation learning tasks [83].